\documentclass{article}

\usepackage[utf8]{inputenc}
\usepackage[T1]{fontenc}
\usepackage{amsmath,amssymb,amsfonts}
\usepackage{graphicx}
\usepackage{booktabs}
\usepackage{hyperref}
\usepackage{cleveref}
\usepackage{algorithm}
\usepackage{algorithmic}
\usepackage{pifont}

\usepackage{tabularx}
\usepackage{array}
\usepackage{xcolor}
\usepackage{natbib}
\usepackage{tikz}
\usepackage{pgfplots}
\pgfplotsset{compat=1.18}
\usetikzlibrary{patterns,patterns.meta,arrows.meta,positioning,calc,backgrounds,shapes.geometric,decorations.pathreplacing,fit}
\usepgfplotslibrary{fillbetween, groupplots}
\definecolor{agonblue}{HTML}{6366F1}   
\definecolor{agongold}{HTML}{F59E0B}   
\definecolor{agongreen}{HTML}{10B981}  
\definecolor{grpogray}{HTML}{94A3B8}   
\definecolor{deadred}{HTML}{DC2626}    
\definecolor{inkblue}{HTML}{1E293B}    
\pgfplotsset{
  agonaxis/.style={
    width=0.6\linewidth, height=4.5cm,
    axis line style={gray!45}, tick style={gray!45},
    grid=major, grid style={gray!12},
    tick label style={font=\footnotesize}, label style={font=\footnotesize},
    legend style={font=\footnotesize, draw=none, fill=none, row sep=1pt},
    title style={font=\small\bfseries, inkblue},
    every axis plot/.append style={line width=1.1pt},
  },
}
\tikzset{
  ncard/.style={rounded corners=4pt, draw=#1!70!black, fill=#1!10, line width=0.8pt,
                inner sep=4pt, align=center, font=\footnotesize},
  glow/.style={preaction={draw=#1!35, line width=3.5pt, opacity=0.5}},
}
\usepackage[margin=1in]{geometry}

\newcommand{\method}{Agon}

\title{\method{}: Competitive Cross-Model RL\\with Implicit Rival Grading of Reasoning}
\author{Vladislav Beliaev\\
Independent Researcher\\
\texttt{belyaev.vladislav.nw@gmail.com}\\
\href{https://thinkdense.ai}{\texttt{thinkdense.ai}}}
\date{}

\begin{document}

\maketitle

\begin{abstract}
Reinforcement learning from verifiable rewards (e.g.\ GRPO) is the engine behind today's reasoning models, yet it grades only the final answer. On hard problems this trains models to write more rather than to think better, since the trace itself is never graded and no label for good thinking exists. We introduce \textbf{\method{}}, which makes two competing models each other's graders. Both attempt the same problem; in alternating roles, one drafts a solution and the other reads it while solving, and each is rewarded for out-solving the other. To win, a model must out-reason a rival that has seen its work, so reasoning is judged implicitly during training, with no process labels and no reward model. Because both models are optimized, each faces a progressively stronger rival, which single-model RL cannot provide. The two need only be comparably strong and behaviorally different. At inference the pair deploys as it trains, a two-stage cascade in which one model drafts and the other answers after reading the draft. On the hard split of DeepMath with Qwen3, this doubles GRPO's pass@1, roughly eight times the gain of an untrained Mixture-of-Agents pass over the same base. The ordering replicates on competitive-programming code and across model families (Qwen3.5, Gemma~4). For now the models talk in text; the next step is to let them reason together in latent space.
\end{abstract}

\section{Introduction}
\label{sec:intro}

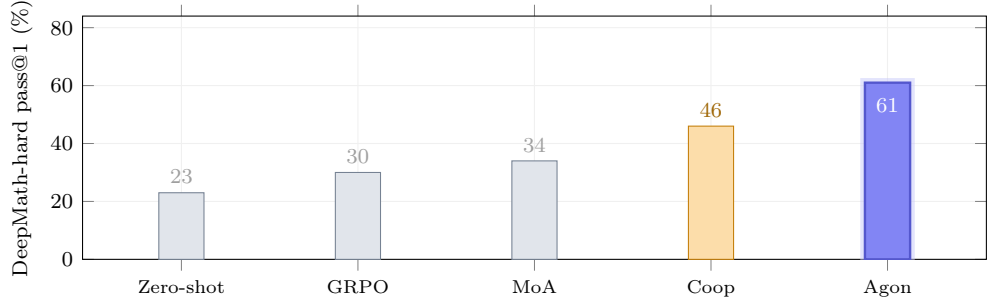
\begin{figure}[t]
\centering
\begin{tikzpicture}
\begin{axis}[
  width=0.82\linewidth, height=4.8cm,
  ybar, bar width=17pt,
  ymin=0, ymax=84,
  ylabel={DeepMath-hard pass@1 (\%)}, ylabel style={font=\footnotesize},
  symbolic x coords={Zero-shot, GRPO, MoA, Coop, Agon},
  xtick={Zero-shot, GRPO, MoA, Coop, Agon}, xticklabel style={font=\scriptsize},
  ytick={0,20,40,60,80}, tick label style={font=\footnotesize},
  grid=major, grid style={gray!12},
  nodes near coords, nodes near coords style={font=\footnotesize\bfseries},
  enlarge x limits=0.14,
]
\addplot[bar shift=0pt, draw=grpogray!80!black, fill=grpogray!28, nodes near coords style={gray!75}]
  coordinates {(Zero-shot,23) (GRPO,30) (MoA,34)};
\addplot[bar shift=0pt, draw=agongold!80!black, fill=agongold!35,
  nodes near coords style={agongold!65!black}]
  coordinates {(Coop,46)};
\addplot[bar shift=0pt, draw=agonblue!85!black, fill=agonblue!80, line width=0.9pt, glow=agonblue,
  nodes near coords style={white, anchor=north, yshift=-2pt, font=\footnotesize\bfseries}]
  coordinates {(Agon,61)};
\end{axis}
\end{tikzpicture}
\caption{From answer-grading to rival-grading. On a hard-math held-out set, vanilla GRPO lifts a Qwen3 model only modestly over zero-shot, and an untrained Mixture-of-Agents pass (MoA, no train), two models cross-refining \emph{at inference}, adds a small bump. Training the same two-model setup competitively (\method{}) reaches roughly $2\times$ GRPO's and $1.8\times$ the untrained MoA's pass@1 at the same two-pass inference budget (\Cref{tab:main}). Intermediate ablations appear in \Cref{sec:experiments}.}
\label{fig:teaser}
\end{figure}

Reinforcement learning from verifiable rewards has become the standard tool for sharpening the reasoning of large language models (LLMs) on tasks such as mathematics and code~\citep{guo2025deepseek,shao2024deepseekmath}. The dominant recipe is on-policy and outcome-based: sample a group of rollouts from the current policy, score each by whether its \emph{final answer} is correct, and update with group-relative advantages, as in GRPO~\citep{shao2024deepseekmath}. The reward is attached to the answer; the chain of thought that produced it is never graded. On problems the model can already partly solve this is benign, but on hard problems it creates a strong incentive to write more: a longer chain affords more chances to stumble onto the right answer, so additional text is the cheapest way to raise expected reward. The trace inflates with hedging and backtracking (``hmm,'' ``wait,'' ``let me reconsider''), and accuracy grows far slower than length (accuracy rises only modestly while length inflates by an order of magnitude~\citep{aggarwal2025l1}). The model ends up producing more reasoning per problem without producing better reasoning per token.

The obvious fix, grading the reasoning directly, is intractable. There is no ground truth for ``good thinking'': no label says which step was the insight and which was filler, and a learned process reward model is expensive, brittle, and itself unverifiable~\citep{lightman2023letsverify}. So the trace stays unscored, and the length pathology persists. We ask instead: \emph{can a second model supply the missing signal?} If a different policy attempts the same problem and we reward each model for \emph{out-reasoning the other}, then the trace is graded implicitly by the rival, without any process labels. A step that leads to a win is reinforced; filler that the opponent exploits is punished.

The grader must be a \emph{different} model, and the game must be \emph{competitive}. Post-training RL is, by construction, self-improvement: the policy is optimized on signal it generates itself, which reinforces the very blind spots that created its errors: a model auditing its own work tends to plateau~\citep{huang2024large}. A second model with different failure modes breaks this closed loop. But merely \emph{having} two models is not enough. Naively averaging or letting models agree, as in self-consistency and mixture-of-agents~\citep{wang2023selfconsistency,wang2024moa}, tends to regress toward consensus and can \emph{dilute} quality when the models differ in strength~\citep{li2025moa}, while two near-identical models gain little from one another. The combination that works is a pair of comparable strength hypothesized to have different blind spots, placed under competitive pressure: each model must not only solve the problem but beat a peer that sees its work, so the objective rewards out-reasoning the peer rather than agreeing with it. Under this pressure the pair co-improves: each win raises the bar for the other on the next step, and both models end up past the ceiling where a single self-graded policy plateaus.

We instantiate this as \textbf{\method{}} (Greek \emph{ag\=on}, contest). Two distinct policies are trained head-to-head: a \emph{drafter} produces candidate solutions; a \emph{challenger} reads an opponent's worked solution and is rewarded for solving correctly \emph{and} beating it; roles rotate every step so both models learn both roles, and at inference the pair deploys the same way, one model drafting and the other answering after reading the draft. In principle the recipe applies to any pair of comparably strong models; for efficiency we realize the pair as two low-rank adapters over one frozen base (${\sim}2\%$ memory overhead, not $2\times$), kept divergent by different initialization and update streams. In this work the exchange happens in text: we first establish that the cross-model signal is useful before optimizing the channel (latent-space exchange; \Cref{sec:conclusion}). We test two claims under a matched generation budget: that cross-model information exchange raises accuracy (the \emph{exchange claim}), and that adversarial pressure raises it further than cooperation (the \emph{competition claim}), using a $2\times2$ design that crosses competition with exchange (one cell of which is provably inert under group-relative normalization; \Cref{sec:reward}), plus a single-model self-refinement control, a zero-training orchestration control, and a budget-matched GRPO self-cascade control.

\paragraph{Contributions.}
\begin{itemize}
\item We reframe \emph{reasoning quality as an implicit reward} that a competing peer model can supply, turning the unlabeled ``good thinking'' problem into a head-to-head game played during RL (\Cref{sec:method}).
\item We propose \method{}: a recipe for \emph{jointly training two distinct policies} against each other with competitive GRPO, via draft-and-challenge rollouts, per-step role rotation, and a competitive reward (correctness plus a conversion bonus for out-solving the opponent). It runs on a stock GRPO trainer with no changes to the optimization core; a shared-base dual-adapter instantiation makes the pair nearly as cheap as a single model (\Cref{sec:method}).
\item We lay out a $2\times2$ matrix (competition $\times$ information exchange) with self-refinement and zero-training controls (\Cref{sec:experiments}). One cell, competition without exchange, is inert by construction under group-relative normalization (\Cref{sec:reward}), so the competition claim rests on the adversarial-vs-cooperative comparison; the fully isolated cell (per-rollout hidden drafts) is left for future work.
\item Under a matched generation budget on hard math, competition $+$ exchange beats cooperation, self-refinement, and vanilla GRPO on pass@1, with the largest gains on the smallest models. At the same doubled inference budget, the trained cascade nearly doubles the accuracy of an untrained Mixture-of-Agents pass over the same base, which itself barely improves on GRPO.
\end{itemize}

\section{Related Work}
\label{sec:related}

\paragraph{Outcome-only RL and the length pathology.} GRPO~\citep{shao2024deepseekmath} and the R1 recipe~\citep{guo2025deepseek} reward only the verified final answer. A well-documented side effect is uncontrolled growth of trace length (``overthinking''), with accuracy rising far more slowly than tokens~\citep{aggarwal2025l1,chen2024overthink}. Remedies act on the symptom: explicit length penalties or length-controlled objectives~\citep{aggarwal2025l1}, and dynamic sampling that discards degenerate all-correct/all-wrong groups~\citep{yu2025dapo}. None grades the \emph{content} of the trace, because step-level labels do not exist and learned process reward models are costly and themselves unverifiable~\citep{lightman2023letsverify}. \method{} leaves the verifier untouched and instead derives a content signal from a competing peer.

\paragraph{Self-play and self-improvement.} A model can improve against versions of itself: SPIN~\citep{chen2024spin} plays the current policy against its own past generations; Self-Rewarding LMs~\citep{yuan2024selfreward} use the model as its own judge; and self-play reasoners such as Absolute Zero~\citep{zhao2025absolutezero} and R-Zero~\citep{huang2025rzero} co-evolve a proposer and a solver instantiated from one model. These methods are powerful but share a structural ceiling: a policy optimized on its own signal reinforces the blind spots that caused its errors, and self-correction without external feedback is unreliable~\citep{huang2024large}. \method{} differs by training \emph{two distinct} policies with different initialization and update streams, so the grader is a genuinely different agent rather than a copy of the graded. Multiagent finetuning~\citep{subramaniam2025multiagent} also diversifies several models from one base via distinct data streams, but toward cooperative ensembling; \method{} maintains divergence for \emph{adversarial grading}, where the distinct rival is what supplies the reward.

\paragraph{Multi-model collaboration, debate, and ensembles.} Combining several models (self-consistency~\citep{wang2023selfconsistency}, multi-agent debate~\citep{du2023debate}, and Mixture-of-Agents~\citep{wang2024moa}) can improve robustness, but the aggregation is \emph{cooperative}: it pushes toward consensus, which dilutes quality when members differ in strength~\citep{li2025moa}, and yields little when they are near-identical. These are inference-time procedures over frozen models. \method{} instead applies cross-model pressure \emph{during} RL and makes it competitive: a model is rewarded for out-solving a peer that has seen its work, so the objective pays for beating the peer, not for matching it. We use a cooperative variant only as an ablation to isolate the value of competition, and include a two-agent MoA layer as a zero-training control; it recovers roughly one-eighth of \method{}'s gain over GRPO (\Cref{tab:main}). Closest in spirit is in-context self-refinement~\citep{madaan2023selfrefine}, where a model revises its \emph{own} draft; \method{} differs in that the draft comes from a \emph{competing, distinct} peer and is selected by outcome under RL, not by self-feedback.

\paragraph{Adversarial and generator--verifier training.} Discriminator/verifier signals have long shaped generators, from GANs~\citep{goodfellow2014gan} to prover--verifier games that improve the legibility of LLM solutions~\citep{kirchner2024prover} and adversarial games over language~\citep{cheng2024spag}. Debate~\citep{irving2018debate} first framed a judged competition between two agents as a scalable proxy for supervising reasoning itself. \method{} is symmetric and rotational rather than role-fixed: both models alternate between drafting and challenging, so neither degenerates into a pure critic, and both improve as solvers; the judge is a ground-truth verifier rather than a learned or human one. The eventual move from text to latent-space exchange connects to work on inter-model communication in representation space~\citep{esperantix2025}, which we leave as future work.

\section{Preliminaries: GRPO and the Unscored Trace}
\label{sec:prelim}
For a problem $x$ with a verifier $r(x,y)\in\{0,1\}$ that checks only the final answer of a completion $y=(\text{trace},\text{answer})$, GRPO~\citep{shao2024deepseekmath} samples a group of $G$ rollouts $y_1,\dots,y_G\sim\pi_\theta(\cdot\mid x)$ and forms the group-relative advantage
\begin{equation}
A_i \;=\; \frac{r(x,y_i)-\mu}{\sigma},\qquad \mu=\tfrac{1}{G}\textstyle\sum_j r(x,y_j),\quad \sigma=\mathrm{std}_j\,r(x,y_j),
\label{eq:grpo-adv}
\end{equation}
updating $\pi_\theta$ to raise the log-probability of above-average rollouts, with the usual KL regularization toward a reference (the base) omitted here for brevity; groups with zero reward variance (all-correct or all-wrong, where $\sigma=0$) carry no signal and are dropped. The signal is entirely a function of the \emph{answer}: two completions with the same answer receive the same advantage regardless of how they reasoned.

\paragraph{The unscored trace.} On hard problems the per-problem solve rate $\mathbb{E}_{y\sim\pi_\theta}[r(x,y)]$ is small, and the policy raises expected reward most cheaply by increasing the number of distinct attempts inside one trace: each restart, case split, or ``let me reconsider'' is another chance at the answer. Because the trace is never scored, this padding is not penalized and is in fact reinforced through any correct rollout that contains it. The result is the empirically observed regime where length grows much faster than accuracy~\citep{aggarwal2025l1,chen2024overthink}; in our own runs, GRPO training grows the mean trace from the zero-shot $6.1$k to $8.1$k tokens for $+7$\,pp accuracy (\Cref{tab:main}). The policy increases the quantity of reasoning while its density, the useful signal per token, stays flat. The bottleneck, then, is not quantity. The prevailing axes of scale (data, parameters, context, and sampled tokens) all add more reasoning; \method{} instead targets a different axis, signal per token, by making low-density traces lose the competition.

\paragraph{The missing grader.} Closing this gap requires a reward that distinguishes a concise, sound trace from a long, lucky one, i.e.\ a signal on the reasoning itself. Verifiers do not provide it, and process reward models require labels we do not have~\citep{lightman2023letsverify}. \method{} obtains the signal \emph{relationally}: a peer policy attempts the same $x$, reads the other's worked solution (the post-reasoning summary, not the raw trace; \Cref{sec:step}), and the reward depends on \emph{which model out-reasons the other}. A trace whose flaws the opponent exploits (a sign error, a dead end it avoids) loses; a trace that leads to a win is reinforced. The grader is thus a second model, and, because both models are optimized, the bar rises over training. We formalize this next.

\section{Method: \method{}}
\label{sec:method}
\method{} trains two policies, $A$ and $B$, head-to-head on the same problems. The design has four parts: a draft-and-challenge rollout that routes one model's reasoning into the other's context (\Cref{sec:step}); a competitive reward that grades the challenger against the drafter (\Cref{sec:reward}); a compute-parity accounting so all comparisons are fair (\Cref{sec:compute}); and an efficient instantiation of the two policies as divergent adapters over a shared base (\Cref{sec:arch}).

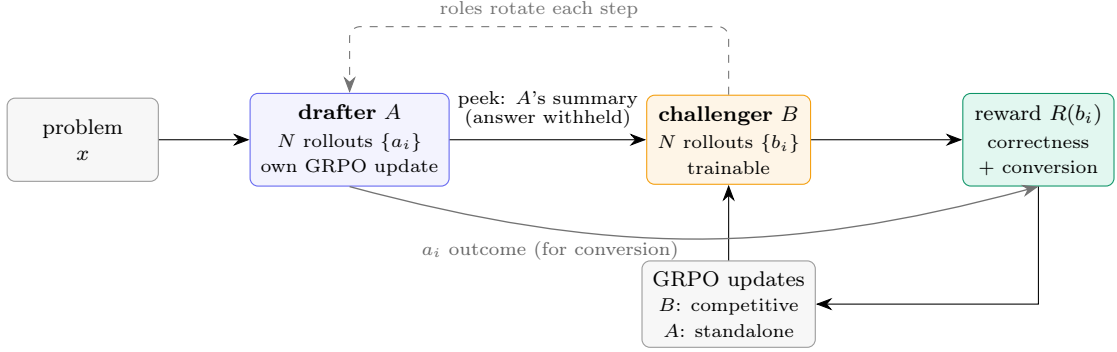
\begin{figure}[t]
\centering
\begin{tikzpicture}[
  font=\footnotesize, node distance=16mm,
  box/.style={rounded corners=3pt, draw, minimum height=11mm, minimum width=20mm, inner sep=4pt, align=center},
  Abox/.style={box, draw=agonblue, fill=agonblue!8},
  Bbox/.style={box, draw=agongold, fill=agongold!10},
  Rbox/.style={box, draw=agongreen!80!black, fill=agongreen!12},
  >={Stealth[length=2.2mm]},
]
\node[box, draw=black!40, fill=black!3] (prob) {problem\\$x$};
\node[Abox, right=12mm of prob] (draft) {\textbf{drafter} $A$\\[1pt]\scriptsize $N$ rollouts $\{a_i\}$\\\scriptsize own GRPO update};
\node[Bbox, right=26mm of draft] (chal) {\textbf{challenger} $B$\\[1pt]\scriptsize $N$ rollouts $\{b_i\}$\\\scriptsize trainable};
\node[Rbox, right=20mm of chal] (rew) {reward $R(b_i)$\\[1pt]\scriptsize correctness\\\scriptsize $+$ conversion};
\node[box, draw=black!40, fill=black!3, below=10mm of chal] (grpo) {GRPO updates\\\scriptsize $B$: competitive\\\scriptsize $A$: standalone};

\draw[->] (prob) -- (draft);
\draw[->] (draft) -- node[above=1pt,midway,font=\scriptsize,align=center]{peek: $A$'s summary\\[-1pt](answer withheld)} (chal);
\draw[->] (chal) -- (rew);
\draw[->] (rew) |- (grpo);
\draw[->] (grpo) -- (chal);
\draw[->, black!55, line width=0.5pt] (draft.south) to[out=-15, in=-165]
  node[below=1pt, pos=0.28, font=\scriptsize, black!60]{$a_i$ outcome (for conversion)} (rew.south);
\draw[->, dashed, black!55, rounded corners=4pt]
  (chal.north) -- ++(0,9mm) -| node[pos=0.25, above, font=\scriptsize]{roles rotate each step} (draft.north);
\end{tikzpicture}
\caption{One \method{} step. The drafter $A$ produces $N$ solutions from the plain prompt and receives a vanilla GRPO update on them (its standalone stream); the challenger $B$ reads each opponent's \emph{post-reasoning solution summary} in-context (final answer withheld), produces one paired rollout per opponent ($N$ total), and receives the competitive reward, which uses the paired draft's outcome for the conversion bonus. Both adapters are updated every step. Roles rotate every optimizer step, so both models train in both streams. $A$ and $B$ are two LoRA adapters over one frozen base.}
\label{fig:step}
\end{figure}

\subsection{Why a second model}
\label{sec:why}
\begin{figure}[t]
\centering
\begin{tikzpicture}[font=\footnotesize, >={Stealth[length=2.2mm]},
   mdl/.style={circle, draw, minimum size=11mm, font=\bfseries, line width=1pt}]
 \node[font=\scriptsize\bfseries, deadred!80!black] at (1.25,1.95) {self-play: closed loop};
 \node[font=\scriptsize\bfseries, agongreen!50!black] at (6.1,1.95) {\method{}: open loop};
 \node[font=\scriptsize\bfseries, gray!55] at (3.4,0.65) {vs};
 \node[mdl, draw=deadred!70!black, fill=deadred!8] (a1) at (1.25,0.65) {A};
 \draw[->, deadred!70, line width=1.2pt] ([shift=(65:8.5mm)]a1.center) arc (65:395:8.5mm);
 \node[mdl, draw=agonblue!70!black, fill=agonblue!10] (a2) at (5.4,0.65) {A};
 \node[mdl, draw=agongold!75!black, fill=agongold!16] (b2) at (6.8,0.65) {B};
 \draw[->, agonblue!80, line width=1.1pt] ([yshift=2.2pt]a2.east) -- ([yshift=2.2pt]b2.west);
 \draw[->, agongold!85, line width=1.1pt] ([yshift=-2.2pt]b2.west) -- ([yshift=-2.2pt]a2.east);
 \node[font=\scriptsize, align=center, deadred!80!black] at (1.25,-0.75) {audits itself with its own bias\\$\to$ plateau};
 \node[font=\scriptsize, align=center, agongreen!50!black] at (6.1,-0.75) {graded by different blind spots\\$\to$ the grader improves too};
\end{tikzpicture}
\caption{Why a second model. \textbf{Left:} a single policy optimized on its own rollouts re-audits with the same biases that caused its errors, so self-play saturates~\citep{huang2024large}. \textbf{Right:} \method{} pairs two divergent policies; each grades the other with different blind spots, each is rewarded for winning, and because both are optimized, the grader improves with the graded.}
\label{fig:loop}
\end{figure}
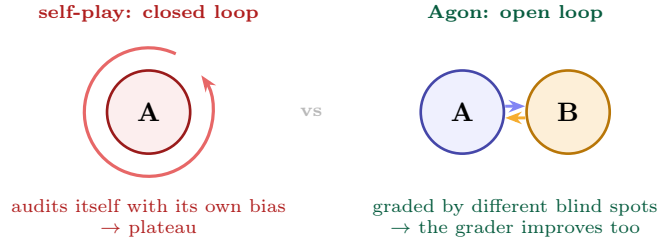

Post-training RL is self-improvement: the gradient in \Cref{eq:grpo-adv} is computed from rollouts the policy itself produced, so it amplifies whatever the policy already does, including the systematic errors that define its blind spots. A model checking its own work inherits the bias that produced the mistake, which is why unaided self-correction often fails to add signal~\citep{huang2024large}; we expect the same closed loop to cap self-play, however many rounds it runs (illustrated in \Cref{fig:loop}, left). A \emph{different} model is outside this loop: a peer hypothesized to have different failure modes audits what the policy cannot see in itself. In principle the peer could come from a different model family: the strongest models, open ones included, now cluster within a few points of one another on standard reasoning benchmarks~\citep{aiindex2026}, so pairs of matched strength are easy to find; whether such pairs also differ enough in behavior is an empirical question we leave to future work. Our instantiation instead engineers the divergence structurally from one base (\Cref{sec:arch}).

Pairing helps only under two conditions. The models must be of \emph{comparable strength} (otherwise, we expect, the weaker simply imitates the stronger and the game collapses into distillation; we do not ablate the strength gap), and they must have \emph{different failure modes}, so that one can catch what the other misses (we make this concrete in \Cref{sec:arch}, \Cref{fig:diverge}). We therefore keep the pair matched in capacity but divergent in behavior, and we make the interaction \emph{competitive} so the objective rewards beating the peer rather than agreeing with it.

\subsection{Instantiating the two policies}
\label{sec:arch}
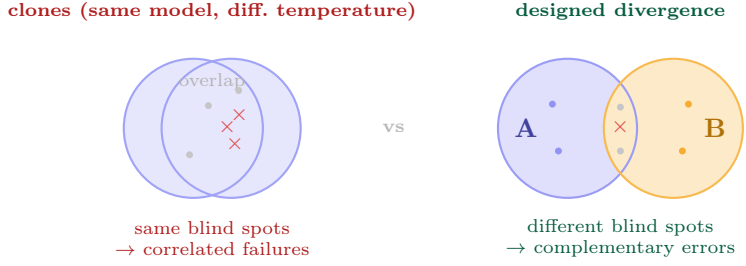
\begin{figure}[t]
\centering
\begin{tikzpicture}[font=\footnotesize]
 \node[font=\scriptsize\bfseries, deadred!80!black] at (1.0,1.55) {clones (same model, diff.\ temperature)};
 \fill[agonblue!16] (0.75,0) circle (0.92);
 \fill[agonblue!16] (1.25,0) circle (0.92);
 \draw[agonblue!60, line width=0.8pt] (0.75,0) circle (0.92);
 \draw[agonblue!60, line width=0.8pt] (1.25,0) circle (0.92);
 \node[font=\scriptsize, gray!60] at (1.0,0.62) {overlap};
 \fill[gray!45] (0.7,-0.35) circle (1.3pt);
 \fill[gray!45] (0.95,0.3) circle (1.3pt);
 \fill[gray!45] (1.35,0.5) circle (1.3pt);
 \node[deadred!85, font=\footnotesize] at (1.2,0.02) {$\times$};
 \node[deadred!85, font=\footnotesize] at (1.36,0.18) {$\times$};
 \node[deadred!85, font=\footnotesize] at (1.3,-0.2) {$\times$};
 \node[font=\scriptsize, deadred!80!black, align=center] at (1.0,-1.45) {same blind spots\\$\to$ correlated failures};
 \node[font=\scriptsize\bfseries, gray!55] at (3.4,0) {vs};
 \node[font=\scriptsize\bfseries, agongreen!50!black] at (6.4,1.55) {designed divergence};
 \begin{scope}[xshift=5.4cm]
   \fill[agonblue!20] (0.3,0) circle (0.92);
   \fill[agongold!22] (1.7,0) circle (0.92);
   \draw[agonblue!65, line width=0.8pt] (0.3,0) circle (0.92);
   \draw[agongold!72, line width=0.8pt] (1.7,0) circle (0.92);
   \node[agonblue!65!black, font=\bfseries] at (-0.25,0) {A};
   \node[agongold!72!black, font=\bfseries] at (2.25,0) {B};
   \fill[agonblue!70] (0.1,0.32) circle (1.3pt);
   \fill[agonblue!70] (0.18,-0.3) circle (1.3pt);
   \fill[agongold!85] (1.9,0.32) circle (1.3pt);
   \fill[agongold!85] (1.82,-0.3) circle (1.3pt);
   \fill[gray!45] (1.0,0.28) circle (1.3pt);
   \fill[gray!45] (1.0,-0.3) circle (1.3pt);
   \node[deadred!85, font=\footnotesize] at (1.0,0.02) {$\times$};
 \end{scope}
 \node[font=\scriptsize, agongreen!50!black, align=center] at (6.4,-1.45) {different blind spots\\$\to$ complementary errors};
\end{tikzpicture}
\caption{Designed divergence (schematic of the target regime, not a measurement). Two copies of one model (left) cover the same problems and fail on the same ones, so a second copy adds little. A pair that is equally strong but behaviorally different (right) has complementary error profiles, which is what makes cross-model grading pay off. Our dual-adapter instantiation starts near the left picture (same base, noise-level adapter difference) and relies on divergent initialization and rotated update streams to move toward the right during training; we do not measure how far it gets.}
\label{fig:diverge}
\end{figure}

\method{} itself only requires two policies of matched strength and different blind spots; how they are parameterized is an implementation choice. We instantiate both as LoRA adapters~\citep{hu2022lora} over a single frozen base model: $\pi_A=\mathrm{base}\!+\!\Delta_A$ and $\pi_B=\mathrm{base}\!+\!\Delta_B$. The second policy costs one extra rank-16 adapter (${\sim}10$M parameters, ${\sim}2\%$ of the base, plus its optimizer state) rather than a full second model, and the shared base keeps the two policies in compatible representation spaces, useful for the eventual latent-space exchange (\Cref{sec:conclusion}). Two models alone are not enough, though: the useful regime is the one motivated in \Cref{sec:why}, \emph{equal strength with hypothesized different blind spots} (\Cref{fig:diverge}). We maintain divergence by (i) different initialization ($B$ starts with small Gaussian noise around the zero matrix; $A$ uses the standard LoRA zero-init) and (ii) different update streams induced by role rotation (\Cref{sec:step}): on any given problem one adapter drafts while the other challenges, so the two never receive the same gradient. All reported numbers use the same base checkpoint for both adapters. We stress that this instantiation is \emph{minimally} divergent: at initialization the two policies differ only by adapter noise, so any complementarity must emerge from the diverging update streams during training rather than from pre-existing blind-spot differences. Different blind spots are therefore a hypothesis about the trained pair, not a measured property. The gap between cooperative exchange and single-adapter self-refinement (46 vs 32; \Cref{tab:main}) shows that training to read a \emph{peer's} draft beats training to read one's own, but the comparison varies more than one factor (trainable adapter capacity, whose draft is read, the refiner prompt), so it does not isolate blind-spot divergence as the cause.

\subsection{The draft-and-challenge step}
\label{sec:step}
Each optimizer step (\Cref{fig:step}, \Cref{alg:agon}) designates a \emph{drafter} and a \emph{challenger}. The drafter generates $N$ rollouts $\{a_i\}\sim\pi_{\text{draft}}(\cdot\mid x)$ from the plain problem prompt and receives a vanilla GRPO update on them (correctness and format only, no opponent term); this standalone stream trains each adapter to solve from scratch, which is the first stage of deployment (\Cref{sec:compute}). The challenger then conditions on its paired opponent's \emph{solution summary}: the answer section the model writes after its private reasoning block, which recaps the derivation. The raw thinking trace is dropped, since the summary carries the distilled solution at a fraction of the tokens (cheaper to prefill, and free of the exploration noise that dominates long traces), and the \emph{final answer is withheld} as a precaution against copying (\Cref{fig:peek}), though a worked summary often implies its answer anyway. The summary is inserted as an attempt that ``may or may not be correct,'' and the challenger generates one rollout per pair, $b_i\sim\pi_{\text{chal}}(\cdot\mid x, a_i)$, forming a group of $N$. The challenger receives the competitive gradient; each stream's GRPO advantage is computed within its own group of $N$. Because each $b_i$ sees a different opponent $a_i$, the group spans varied opponent contexts for the same $x$; this is what gives the conversion bonus within-group variance. \textbf{Roles rotate every step}: on even steps $A$ drafts and $B$ challenges, on odd steps the reverse. Without rotation, each adapter receives only one kind of gradient, one only ever solving from scratch and the other only ever solving against an opponent; the pair still trains, but ends up weaker (52 vs 61 pass@1; \Cref{tab:ablations}). With rotation, both adapters alternate between the two streams every step and co-improve. The challenger prompt instructs the model not to trust or paraphrase the opponent but to verify it and solve independently, winning by being correct when the opponent is wrong; the prompt is identical across the coop and adv reward stacks (\Cref{sec:reward}), so they differ only in the reward.

\begin{figure}[t]
\centering
\begin{tikzpicture}[font=\footnotesize, >={Stealth[length=2.4mm]}]
  \fill[agonblue!9]  (1.6,5.1) -- (0.1,1.1) -- (3.1,1.1) -- cycle;
  \draw[agonblue!55, line width=0.9pt] (1.6,5.1) -- (0.1,1.1) -- (3.1,1.1) -- cycle;
  \draw[agonblue!22, line width=0.4pt, dotted] (1.6,4.7) -- (1.6,1.35);
  \node[agonblue!70!black, font=\bfseries] at (1.6,5.4) {A};
  \node[gray!50, font=\scriptsize] at (1.6,1.6) {reasoning};
  \fill[agongold!10] (8.4,5.1) -- (6.9,1.1) -- (9.9,1.1) -- cycle;
  \draw[agongold!65, line width=0.9pt] (8.4,5.1) -- (6.9,1.1) -- (9.9,1.1) -- cycle;
  \draw[agongold!28, line width=0.4pt, dotted] (8.4,4.7) -- (8.4,1.35);
  \node[agongold!75!black, font=\bfseries] at (8.4,5.4) {B};
  \node[gray!50, font=\scriptsize] at (8.4,1.6) {reasoning};
  \node[gray!60, font=\scriptsize\itshape] at (5.0,5.05) {each peeks at the rival's summary};
  \draw[->, agonblue!80, line width=1pt, dash pattern=on 3pt off 2pt] (2.0,4.0) -- (8.0,4.0);
  \node[agonblue!70!black, font=\scriptsize, anchor=south] at (5.0,4.06) {spots a sign error in eq.\ 3};
  \draw[->, agongold!85, line width=1pt, dash pattern=on 3pt off 2pt] (7.6,3.0) -- (2.39,3.0);
  \node[agongold!75!black, font=\scriptsize, anchor=south] at (5.0,3.06) {skips a failed hypothesis};
  \draw[->, agonblue!80, line width=1pt, dash pattern=on 3pt off 2pt] (2.84,1.95) -- (7.22,1.95);
  \node[agonblue!70!black, font=\scriptsize, anchor=south] at (5.0,2.01) {borrows the angle symmetry};
  \node[ncard=agonblue, anchor=north, minimum width=2.1cm] (aa) at (1.6,0.85) {$A$'s answer};
  \node[ncard=agongold, anchor=north, minimum width=2.1cm] (bb) at (8.4,0.85) {$B$'s answer};
  \node[ncard=agongreen, anchor=north, font=\scriptsize\bfseries] (q) at (5.0,0.85) {who out-scores?};
  \draw[->, gray!55] (aa.east) -- (q.west);
  \draw[->, gray!55] (bb.west) -- (q.east);
  \node[anchor=north, gray!70, font=\scriptsize] at (5.0,-0.35) {reward $=$ correctness $+$ bonus for out-solving the rival};
\end{tikzpicture}
\caption{Peek and outscore. A conceptual view aggregated over one rotation cycle: within a single step the exchange is one-directional (\Cref{fig:step}); roles swap on the next step, so each model both reads and is read, on different problems. While solving, the challenger reads the rival's \emph{worked solution} (the summary it writes after its private reasoning, final answer withheld) and writes against it, catching its errors (a sign error in eq.~3), pruning its dead ends, and borrowing its good moves; the annotations are what the reader of the summary exploits, not messages exchanged between the models. Neither solution is labeled, yet by rewarding whoever out-solves the other, the rival's reasoning is graded implicitly, a signal that outcome-only RL does not use.}
\label{fig:peek}
\end{figure}
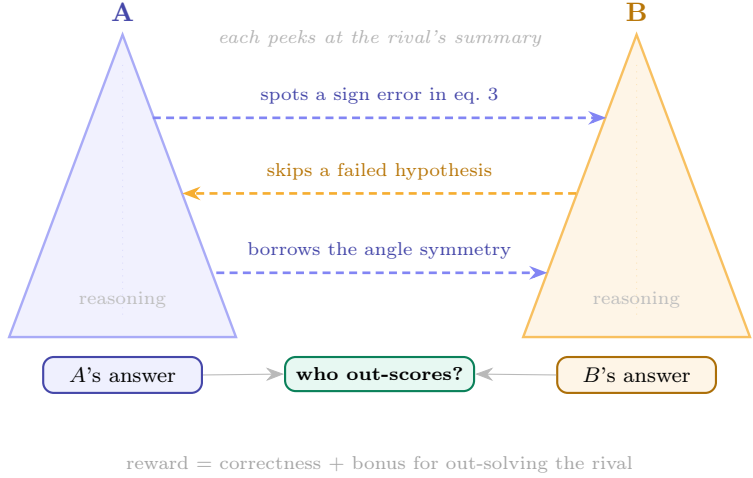

\begin{figure}[t]
\centering
\footnotesize
\fcolorbox{agonblue!55}{agonblue!4}{\begin{minipage}[t]{0.455\linewidth}
\textbf{\textcolor{agonblue!70!black}{A (drafter)}}\\[2pt]
\texttt{<think>} Okay, so I need to compute this integral. Set $u=1-x^2$, so $du=2x\,dx$. Wait, that's negative? Hmm, that can't be right. Maybe I made a mistake. Let me try integration by parts instead\ldots{} \textcolor{gray!70}{\emph{(continues for ${\sim}1.9$k tokens; never shown to $B$)}} \texttt{</think>}\\[3pt]
\textbf{summary} \textcolor{gray!70}{\emph{(what $B$ sees)}}\textbf{:} Substituting $u=1-x^2$ with $du=2x\,dx$ \textcolor{deadred!80!black}{(sign dropped: $du=-2x\,dx$)} gives $\int_0^1 x\sqrt{1-x^2}\,dx=\tfrac12\int_1^0\!\sqrt{u}\,du$, a negative value\ldots\\[2pt]
\textcolor{gray!70}{\emph{final answer withheld from $B$}}
\end{minipage}}
\hfill
\fcolorbox{agongold!65}{agongold!6}{\begin{minipage}[t]{0.455\linewidth}
\textbf{\textcolor{agongold!75!black}{B (challenger), reads $A$'s summary}}\\[2pt]
\texttt{<think>} The attempt's substitution is right but the sign of $du$ was dropped: $x\,dx=-\tfrac12\,du$, which flips the limits back and gives $+\tfrac13$. The area is positive, no need to re-explore. Check at the boundary: integrand $\ge 0$ on $[0,1]$. \texttt{</think>}\\[2pt]
\texttt{<answer>} $1/3$ \texttt{</answer>}\hfill\textcolor{agongreen!50!black}{\textbf{correct, ${\sim}4\times$ shorter}}
\end{minipage}}
\caption{Illustrative exchange (hand-constructed for clarity). The drafter drops a sign, distrusts its own (correct-magnitude) result, and spirals into re-exploration; its post-reasoning summary recaps the flawed derivation (but does not contain the boxed answer). The challenger sees \emph{only} that summary (raw thinking trace dropped, final answer withheld as a precaution), pinpoints the flawed step, re-derives it, and answers correctly in a fraction of the tokens. The solution summary carries the derivation, which is what the challenger exploits.}
\label{fig:example}
\end{figure}

\subsection{Reward: correctness plus a conversion bonus}
\label{sec:reward}
Let $c(y)=r(x,y)\in\{0,1\}$ be correctness. We report two reward stacks, applied to the challenger rollout $b_i$ given its paired drafter solution $a_i$:
\begin{align}
\textbf{coop:}\quad & R(b_i) \;=\; 2\,c(b_i) \;+\; \lambda\,\phi(b_i), \label{eq:coop}\\
\textbf{adv:}\quad  & R(b_i) \;=\; 2\,c(b_i) \;+\; \underbrace{c(b_i)\,\big(1-c(a_i)\big)}_{\text{conversion bonus}} \;+\; \lambda\,\phi(b_i), \label{eq:adv}
\end{align}
where $\phi(\cdot)$ is a format term (well-formed \texttt{<think>}/\texttt{<answer>} structure) with weight $\lambda=0.5$ (25\% of the correctness term), identical across stacks so it introduces no bias between them. In \textbf{coop} the opponent appears only in the \emph{context}, not in the reward, which isolates the value of information exchange (the exchange claim). In \textbf{adv} the \emph{conversion bonus} $c(b_i)\,(1-c(a_i))$ raises the effective correctness weight from $2$ to $3$ exactly on the paired opponent attempts that \emph{failed}: solving where the opponent could not is worth more, which is the competitive pressure.

The form matters. A na\"ive margin $c(b_i)-c(a_i)$ works poorly: since $c(a_i)$ is \emph{action-independent} (fixed by the pairing $a_i$, not by the challenger's rollout $b_i$), its subtractive term contributes \emph{no expected policy gradient} (exactly in the unnormalized objective; up to the nonlinearity of group standardization after it): it acts as a per-sample baseline, shifting each reward by the drafter's outcome without adding any competitive direction. The tested margin stack, $R=2c(b_i)+\big(c(b_i)-c(a_i)\big)=3c(b_i)-c(a_i)$, makes this concrete: the scalar $3$ cancels under group-standard-deviation normalization and the $-c(a_i)$ term is action-independent (zero expected policy gradient), so in expectation the margin points in the \emph{same direction} as the cooperative gradient. The ablation is consistent with this prediction ($49$ vs coop $46$), though the $3$ pp difference lies within binomial sampling noise. The conversion bonus instead \emph{modulates the correctness weight by opponent difficulty}: because it multiplies the action-dependent $c(b_i)$, the reweighting shifts the gradient itself, and, given the within-group spread of opponent contexts $\{a_i\}$, tilts it toward the opponent attempts that failed. We confirm this empirically (\Cref{tab:ablations}): the margin form barely improves on cooperation, whereas the conversion bonus clearly does. More generally, any reward that is affine in $c(b_i)$ with group-constant coefficients (e.g., a shared opponent draft for the entire challenger group, making $c(a)$ identical for all $N$ rollouts) normalizes to exactly the advantage of the conversion-free reward in the same contexts under group-relative standardization; the conversion term then contributes zero expected policy gradient. Within-group variation in opponent difficulty is what makes the competitive term trainable. In the shared-opponent ablation we therefore fix one opponent draft per problem, \emph{hidden} from the challenger's context and used only for the reward (so the contexts are plain prompts, as in vanilla GRPO); the observed result (32) is consistent with the theoretical prediction that a group-constant conversion bonus adds no directional signal. This departs from the standard GRPO assumption that all rollouts in a group share the same prompt; here each challenger rollout conditions on its own opponent context. The resulting credit assignment is not merely noisier but systematically confounded: a rollout's success probability depends on the quality of the draft it happened to draw (an action-independent draw), so group standardization mixes context difficulty with policy quality, and the conversion bonus, which pays exactly in the harder contexts, partially offsets that difficulty penalty. This admits an alternative reading of the adversarial gain: rather than ``competitive pressure'' in a game-theoretic sense, the bonus acts as difficulty-weighted reward shaping that upweights correct completions in hard contexts. Our experiments do not distinguish these two interpretations; we use the competitive framing for its constructive role in the design and leave the disentanglement to future work.

The resulting reward ladder (adv) is $R-\lambda\phi\in\{0,2,3\}$. The drafter's standalone stream has no opponent, so \Cref{eq:coop} applies as-is.

As an \emph{optional density lever}, studied separately in \Cref{sec:density} and \emph{not} part of the headline recipe, we add a length tiebreak that fires only when both solutions are correct: $R(b_i)\mathrel{+{=}}\gamma\,c(b_i)\,c(a_i)\,\mathbf{1}\!\left[\,|b_i|<|a_i|\,\right]$ with weight $\gamma$, rewarding the challenger for matching the opponent's correctness in \emph{fewer} tokens. Because the term requires $c(b_i)=1$, it never \emph{directly} penalizes correctness, only breaking ties among already-correct traces, so accuracy stays essentially flat while length drops, converting the emergent shortening (\Cref{sec:analysis}) into a directly tunable target.

\begin{algorithm}[t]
\caption{\method{} (one optimizer step)}
\label{alg:agon}
\begin{algorithmic}[1]
\STATE \textbf{input:} batch of problems $\{x\}$; adapters $A,B$; group size $N$; step index $t$; mode $\in\{\text{coop},\text{adv}\}$
\STATE $(\text{draft},\text{chal}) \leftarrow (A,B)$ if $t$ even else $(B,A)$ \hfill// role rotation
\FORALL{$x$ in batch}
  \STATE $\{a_i\}_{i=1}^N \sim \pi_{\text{draft}}(\cdot\mid x)$ \hfill// standalone rollouts, plain prompt
  \STATE $R(a_i) \leftarrow$ reward via \Cref{eq:coop} \hfill// correctness $+$ format, no opponent term
  \STATE $\{b_i\}_{i=1}^N \sim \pi_{\text{chal}}(\cdot\mid x, a_i)$ \hfill// one per pair: challenger $b_i$ sees $a_i$'s summary
  \STATE $R(b_i) \leftarrow$ reward via \Cref{eq:coop}/\Cref{eq:adv} by mode
  \STATE group-relative advantages of $\{R(a_i)\}$ and of $\{R(b_i)\}$ \hfill// \Cref{eq:grpo-adv}, per group
\ENDFOR
\STATE GRPO update on $\pi_{\text{draft}}$ (standalone stream) and on $\pi_{\text{chal}}$ (competitive stream)
\STATE \textbf{return} updated adapters
\end{algorithmic}
\end{algorithm}

\subsection{Compute parity}
\label{sec:compute}
Per step \method{} generates $N$ drafter rollouts plus $N$ challenger rollouts ($2N$), versus $N$ for vanilla GRPO. Both adapters are updated every step, the drafter on its standalone group and the challenger on its opponent-conditioned group; rotation swaps which adapter gets which stream. Our primary invariant is the \emph{generation budget}: every method generates $2N$ rollouts per training problem (vanilla GRPO gets $2N$ instead of its default $N$) under the same completion-length cap and the same single pass over the training problems. Realized token counts then differ only through learned trace length, bounded for every method by the same cap; we do not audit total generated tokens (the drafter-stage length is not reported), but with the rollout count and length cap equalized, the gains are not attributable to a larger generation \emph{budget}. The challenger additionally prefills the opponent summary, which is extra prefill compute we do not equalize (the summary is short relative to a full trace, is amortized cheaply against generation, and is absent at single-adapter inference). At inference \method{} deploys exactly as it trains: a two-stage cascade in which one adapter drafts from the plain prompt and the other reads the draft's summary and produces the final answer. This is two sequential generation passes, a doubled budget relative to a single-pass baseline, so \Cref{tab:main} includes two-pass controls at the same budget (cross-refinement, self-refinement). Reported trace lengths refer to the final (challenger) stage. Both cascade directions ($A{\to}B$ and $B{\to}A$) are evaluated and the better is reported, selected post hoc on the held-out set.

\section{Experiments}
\label{sec:experiments}
\paragraph{Setup.} We train Qwen3 models~\citep{yang2025qwen3} (primarily 0.6B; 1.7B/4B for scaling) on the hard split of DeepMath-103K~\citep{deepmath2025} (difficulty~8, binary-answer questions removed), with a held-out set of $300$ problems. We use group size $G=N=8$ for both streams, LoRA rank $16$, learning rate $5\times10^{-5}$ (linear decay, $10\%$ warmup), one training epoch over the $3{,}000$ training problems, format weight $\lambda=0.5$, length tiebreak $\gamma=0.5$ (used in \Cref{sec:density} only), and a standard format reward (well-formed \texttt{<think>}/\texttt{<answer>}) as $\phi$. Generation uses vLLM~\citep{kwon2023vllm} and training uses TRL~\citep{vonwerra2020trl}. The primary metric is pass@1 on the DeepMath held-out set; we additionally sanity-check on GSM8K~\citep{cobbe2021gsm8k} and MATH-500~\citep{hendrycks2021math} (\Cref{tab:transfer}). For the second-domain study (\Cref{sec:code}) we train on the \textsc{easy}-difficulty subset of CodeContests~\citep{li2022codecontests} with a unit-test verifier ($r=1$ iff all tests pass), under the same protocol and budget. Evaluation protocol: sampled decoding (temperature $0.6$, top-$p$ $0.95$), one rollout per stage, generation budget matched to training ($15$k completion tokens per stage), the answer parsed from the \texttt{<answer>} block; the $300$ held-out problems are difficulty-8 only with binary-answer questions removed, split with a fixed seed before any training. Pair-trained methods are evaluated as the two-stage cascade (\Cref{sec:compute}) in both directions, reporting the better direction; mean trace length reports the final-stage completion. Reported pass@1 values include 95\% Clopper--Pearson binomial confidence intervals over the 300-problem held-out set (\Cref{tab:ci}).

\paragraph{The $2\times2$ design.} We cross two binary factors, \emph{competition} and \emph{information exchange}, giving four cells plus three controls (\Cref{tab:matrix}): self-refinement, an untrained MoA at inference, and a budget-matched GRPO two-pass self-cascade. One caveat is structural: the ``competition without exchange'' cell uses a shared opponent draft, hidden from the challenger and used only for the reward, which by the analysis in \Cref{sec:reward} makes the conversion bonus group-constant and hence gradient-free, so this cell is theoretically equivalent to vanilla GRPO and serves as a consistency check rather than a true isolation of competition. Consequently, the \emph{exchange claim} (exchange helps) is carried by the GRPO-vs-cooperative comparison, and the \emph{competition claim} (competition helps beyond exchange) by the cooperative-vs-adversarial comparison, holding everything else fixed at a matched generation budget; a competition-without-visible-exchange cell with per-rollout hidden drafts is left for future work.

\begin{table}[t]
\centering
\caption{The ablation matrix. Each cell is a training configuration; everything but the two factors is held fixed. Self-refinement is a single-model control for exchange; the untrained MoA at inference is a zero-training Mixture-of-Agents control.}
\label{tab:matrix}
\small
\renewcommand{\arraystretch}{1.25}
\begin{tabularx}{0.9\linewidth}{l >{\raggedright\arraybackslash}X >{\raggedright\arraybackslash}X}
\toprule
 & \textbf{no information exchange} & \textbf{information exchange} \\
\midrule
\textbf{no competition} & vanilla GRPO & cooperative (\Cref{eq:coop}) \\
\textbf{competition}    & competitive, shared opponent (inert by construction; \Cref{sec:reward}) & \textbf{\method{}} (\Cref{eq:adv}) \\
\bottomrule
\end{tabularx}
\\[2pt]
\footnotesize Controls: self-refinement (model sees its \emph{own} prior attempt); untrained MoA at inference (the base model's two \emph{independent} samples critique each other, no training); GRPO two-pass self-cascade (a trained GRPO model refines its own first pass, matching the cascade's inference budget).
\end{table}

\subsection{Main results}

\Cref{tab:main} reports the main comparison on Qwen3-0.6B. All deltas below are read from the \emph{pass@1} column. The trained pair methods and the refinement controls are two-pass at inference; vanilla GRPO and zero-shot are single-pass (see the footnote of \Cref{tab:main}). Information exchange alone (cooperative exchange: $46$) already improves over vanilla GRPO ($30$) and over self-refinement ($32$), supporting the exchange claim; adding competition (\method{}: $61$) improves further over cooperation, supporting the competition claim. The untrained MoA control is a two-agent Mixture-of-Agents layer at inference~\citep{wang2024moa}: each model reads the other's attempt and refines its own. It lifts GRPO by $+4$\,pp ($30\!\to\!34$); \method{} lifts it by $+31$\,pp, roughly eight times the gain of untrained aggregation. We note that the two levers contribute comparably: information exchange adds $+16$\,pp (GRPO\,$\to$\,cooperative) and competition a further $+15$\,pp (cooperative\,$\to$\,\method{}). The dual-adapter architecture alone contributes little: the competitive-shared-opponent variant sits at $32$, barely above single-model GRPO at $30$, so the gain from exchange is attributable to the shared context, not to going from one model to two.

Each method is a single training run; the confidence intervals in \Cref{tab:ci} cover sampling variation over the 300 held-out problems ($\pm{\sim}5.5$\,pp) but not run-to-run training variance. Both headline deltas (exchange: $+16$\,pp; competition: $+15$\,pp) exceed the width of these intervals. The \method{} cascade reaches roughly twice vanilla GRPO's held-out pass@1 while \emph{shortening} final-stage traces; in gain terms, GRPO lifts the base by $+7$\,pp ($23\!\to\!30$) while \method{} lifts it by $+31$\,pp over GRPO ($30\!\to\!61$), and $+26$\,pp over a two-pass GRPO self-cascade at matched inference budget, nearly doubling the untrained MoA cascade.

To decompose the cascade gain, we evaluate the drafter adapter standalone after training (plain prompt, no opponent summary): it reaches 46 pass@1, a +16\,pp lift over vanilla GRPO. This is notable given the budget asymmetry: each adapter drafts on only half the steps with group size $N$, so it receives roughly a quarter of the standalone-rollout gradient volume of the vanilla GRPO baseline (which uses $2N$ on every problem), plus challenger-stream gradients on the alternating steps. Our working hypothesis is that the challenger-stream updates transfer to standalone solving (the adapter is trained to verify and re-derive, skills that a plain prompt also exercises), but we have not tested this with a control (e.g.\ the standalone accuracy of a cooperative-trained or self-refinement adapter), so the attribution of the drafter's +16\,pp to the competitive objective is a hypothesis, not an isolated measurement. The challenger then converts a net 15 of the drafter's 54 failures into correct answers (with possible regressions among the drafter's 46 successes), yielding the final 61. Under this reading, the drafter's lift (+16\,pp) and the challenger's conversion (+15\,pp) contribute comparable shares of the total gain, the former via the hypothesized transfer of challenger-stream skills to standalone solving.

To check that the gains are not an artifact of the training distribution, \Cref{tab:transfer} reports the same checkpoints on GSM8K and MATH-500 without any further tuning: the ordering is preserved with smaller absolute margins than in-distribution ($+13$ and $+19$\,pp over zero-shot, vs $+38$ on DeepMath-hard), as expected for a method whose pressure targets the hard regime.

\begin{table}[t]
\centering
\caption{Transfer sanity checks on Qwen3-0.6B (pass@1, \%; trained rows: single training run, same checkpoints and cascade protocol as \Cref{tab:main}, no further tuning; zero-shot: single evaluation; values rounded to nearest integer).}
\label{tab:transfer}
\small
\begin{tabular}{lcc}
\toprule
Method & GSM8K & MATH-500 \\
\midrule
Zero-shot            & 62 & 45 \\
Vanilla GRPO         & 68 & 52 \\
\method{}            & \textbf{75} & \textbf{64} \\
\bottomrule
\end{tabular}
\end{table}

\begin{table}[t]
\centering
\caption{Main comparison on Qwen3-0.6B, DeepMath-hard held-out (\%), at a matched training generation budget; inference passes per method are noted in the footnote. Single training run per method; 95\% Clopper--Pearson confidence intervals over the 300-problem held-out set are shown in \Cref{tab:ci}. Values rounded to nearest integer.}
\label{tab:main}
\small
\begin{tabular}{lcc}
\toprule
Method & pass@1 & avg.\ len \\
\midrule
Zero-shot                          & 23 & 6.1k \\
Vanilla GRPO (baseline)            & 30 & 8.1k \\
Self-refinement (control)          & 32 & 7.9k \\
GRPO two-pass self-cascade (control) & 35 & 8.0k \\
MoA~\citep{wang2024moa} (no train) & 34 & 6.9k \\
Competitive, shared opponent (single-pass) & 32 & 7.4k \\
Cooperative exchange               & 46 & 5.1k \\
\textbf{\method{} (competition $+$ exchange)}   & \textbf{61} & \textbf{3.5k} \\
\bottomrule
\end{tabular}
\\[2pt]
\footnotesize pass@1 is the primary metric. Pair-trained methods and the refinement controls run two sequential generation passes per problem (for \method{} and cooperative: draft, then challenge reading the draft's summary); both cascade directions are evaluated and the better is reported. Zero-shot and vanilla GRPO are single-pass. avg.\ len is the final-stage completion length only (full cascade uses $\sim$2$\times$ tokens at inference). Zero-shot and the untrained MoA involve no training and are reported as single evaluations. 95\% confidence intervals are Clopper--Pearson binomial over 300 held-out problems.
\end{table}

\begin{table}[t]
\centering
\caption{95\% Clopper--Pearson confidence intervals over the 300-problem held-out set (single training run per method).}
\label{tab:ci}
\small
\begin{tabular}{lcc}
\toprule
Method & pass@1 & 95\% CI \\
\midrule
Zero-shot                          & 23 & [18.4, 28.3] \\
Vanilla GRPO (baseline)            & 30 & [25.0, 35.4] \\
Self-refinement (control)          & 32 & [26.9, 37.5] \\
GRPO two-pass self-cascade         & 35 & [29.7, 40.7] \\
MoA (no train)                     & 34 & [28.8, 39.5] \\
Competitive, shared opponent       & 32 & [26.9, 37.5] \\
Cooperative exchange               & 46 & [40.3, 51.8] \\
\textbf{\method{}}                 & \textbf{61} & \textbf{[55.2, 66.6]} \\
\bottomrule
\end{tabular}
\end{table}

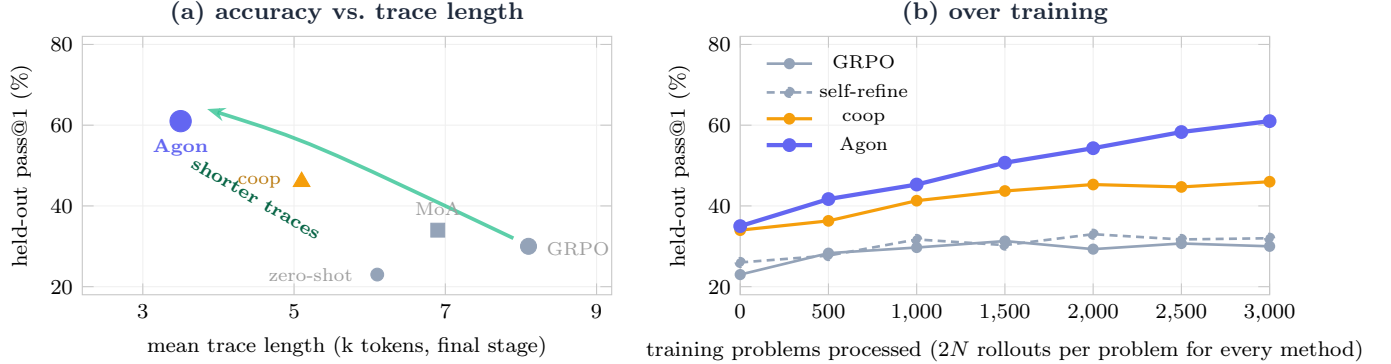
\begin{figure}[t]
\centering
\begin{tikzpicture}
\begin{groupplot}[
  group style={group size=2 by 1, horizontal sep=1.7cm},
  width=0.52\linewidth, height=5.0cm,
  axis line style={gray!45}, tick style={gray!45},
  grid=major, grid style={gray!12},
  tick label style={font=\footnotesize}, label style={font=\footnotesize},
  legend style={font=\scriptsize, draw=none, fill=none, row sep=0.5pt},
  title style={font=\small\bfseries, inkblue},
]
\nextgroupplot[xlabel={mean trace length (k tokens, final stage)}, ylabel={held-out pass@1 (\%)},
  xmin=2.2, xmax=9.2, ymin=18, ymax=82, xtick={3,5,7,9},
  title={(a) accuracy vs.\ trace length}, title style={yshift=-1.2ex}]
\draw[-{Stealth[length=2.4mm]}, agongreen!85, line width=1.4pt, opacity=0.8]
  (axis cs:7.9,32) .. controls (axis cs:5.2,56) .. (axis cs:3.85,64);
\node[font=\scriptsize\bfseries, agongreen!55!black, rotate=-30] at (axis cs:4.5,42) {shorter traces};
\addplot[only marks, mark=*, grpogray, mark size=2.4pt] coordinates {(6.1,23)};
\node[font=\scriptsize, gray!75, anchor=east] at (axis cs:5.9,23) {zero-shot};
\addplot[only marks, mark=*, grpogray, mark size=3pt] coordinates {(8.1,30)};
\node[font=\scriptsize, gray!75, anchor=west] at (axis cs:8.22,29.5) {GRPO};
\addplot[only marks, mark=square*, grpogray, mark size=2.6pt] coordinates {(6.9,34)};
\node[font=\scriptsize, gray!75, anchor=south] at (axis cs:6.9,35.4) {MoA};
\addplot[only marks, mark=triangle*, agongold, mark size=3.6pt] coordinates {(5.1,46)};
\node[font=\scriptsize, agongold!75!black, anchor=east] at (axis cs:4.95,46) {coop};
\addplot[only marks, mark=*, agonblue, mark size=4pt] coordinates {(3.5,61)};
\node[font=\scriptsize\bfseries, agonblue, anchor=north] at (axis cs:3.5,58.8) {\method{}};
\nextgroupplot[xlabel={training problems processed ($2N$ rollouts per problem for every method)}, ylabel={held-out pass@1 (\%)},
  xmin=0, xmax=3000, ymin=18, ymax=82, legend pos=north west,
  xtick={0,500,1000,1500,2000,2500,3000},
  title={(b) over training}, title style={yshift=-1.2ex}]
\addplot[grpogray, line width=1pt, mark=*, mark size=1.6pt] coordinates {(0,23.0)(500,28.3)(1000,29.7)(1500,31.3)(2000,29.3)(2500,30.7)(3000,30.0)};
\addlegendentry{GRPO}
\addplot[grpogray, densely dashed, line width=1pt, mark=*, mark size=1.6pt] coordinates {(0,26)(500,27.7)(1000,31.7)(1500,30.3)(2000,33.0)(2500,31.7)(3000,32.0)};
\addlegendentry{self-refine}
\addplot[agongold, line width=1.2pt, mark=*, mark size=1.6pt] coordinates {(0,34.0)(500,36.3)(1000,41.3)(1500,43.7)(2000,45.3)(2500,44.7)(3000,46.0)};
\addlegendentry{coop}
\addplot[agonblue, line width=1.6pt, mark=*, mark size=1.8pt] coordinates {(0,35.0)(500,41.7)(1000,45.3)(1500,50.7)(2000,54.3)(2500,58.3)(3000,61.0)};
\addlegendentry{\method{}}
\end{groupplot}
\end{tikzpicture}
\caption{\textbf{(a)} \method{} sits in the top-left, higher accuracy at \emph{shorter} traces (lengths are final-stage completions), while GRPO buys length for little accuracy and the untrained MoA adds only a small bump. The arrow is illustrative. \textbf{(b)} Held-out pass@1 at evaluation checkpoints over training. Each point is a single sampled evaluation of a single training run (binomial noise ${\sim}\pm5.5$\,pp at 300 problems); curves are not averaged over seeds. Self-refinement plateaus just above GRPO, cooperative exchange climbs higher, and competition$+$exchange (\method{}) separates clearly. \method{} is still improving at the end of the budget; we stop at the matched budget rather than at convergence.}
\label{fig:results}
\end{figure}

\subsection{Scaling and generality}
We expect smaller models, which live deeper in the hard regime, to benefit most. \Cref{tab:scaling} shows the gain over vanilla GRPO shrinking but remaining positive as the base grows, and the same pattern holding on two other model families, Qwen3.5~\citep{qwen35report} and Gemma~4~\citep{gemma4report}: the delta tracks zero-shot strength, not the family. The \method{} cascade on the 0.6B base (61) exceeds not only zero-shot Qwen3-4B (52) but vanilla GRPO on the 4B base (59), a model ${\sim}7\times$ larger, on the same held-out set. In every case the pair is two adapters over \emph{one} base (our instantiation, not a requirement of the method); pairing across families is future work.

\begin{table}[t]
\centering
\caption{Model scaling and family generality: held-out pass@1 (\%; trained cells: single training run, cascade protocol of \Cref{tab:main}, rounded to the nearest point; zero-shot: single evaluation).}
\label{tab:scaling}
\small
\begin{tabular}{lcccc}
\toprule
Model & Zero-shot & Vanilla GRPO & \method{} & $\Delta$ \\
\midrule
Qwen3-0.6B & 23 & 30 & 61 & \textbf{+31} \\
Qwen3-1.7B & 38 & 46 & 70 & \textbf{+24} \\
Qwen3-4B   & 52 & 59 & 71 & \textbf{+12} \\
\midrule
Qwen3.5-2B & 44 & 50 & 70 & \textbf{+20} \\
Gemma-4-E4B & 50 & 58 & 73 & \textbf{+15} \\
\bottomrule
\end{tabular}
\end{table}

\subsection{A second domain: competitive-programming code}
\label{sec:code}
The recipe is not specific to math; any domain with a clean programmatic verifier qualifies. We repeat the core comparison on Qwen3-1.7B trained on the \textsc{easy}-difficulty subset of CodeContests~\citep{li2022codecontests}, with $r(x,y)=1$ iff the generated program passes all unit tests, a held-out set of $300$ problems, and the same protocol and generation budget. \Cref{tab:code} shows the same ordering as \Cref{tab:main}, exchange beats the baseline and competition beats cooperation, with traces again shortening. The absolute delta is smaller than on math: unit-test rewards are sparser at this scale, and code traces make the opponent prefill longer relative to generation. Across domains the claims should be read as directional rather than magnitude-matched.

\begin{table}[t]
\centering
\caption{Second domain: CodeContests (\textsc{easy} difficulty) on Qwen3-1.7B, held-out pass@1 (\%; single training run; zero-shot: single evaluation; values rounded to nearest integer).}
\label{tab:code}
\small
\begin{tabular}{lcc}
\toprule
Method & pass@1 & avg.\ len \\
\midrule
Zero-shot            & 18 & 5.2k \\
Vanilla GRPO         & 24 & 7.3k \\
Cooperative exchange & 29 & 4.8k \\
\textbf{\method{}}   & \textbf{34} & \textbf{3.9k} \\
\bottomrule
\end{tabular}
\end{table}

\subsection{Ablations}
\label{sec:ablations}
\Cref{tab:ablations} isolates each design choice on Qwen3-0.6B (held-out pass@1). The competition claim is carried by the \emph{competition} row (adv vs coop). The \emph{information exchange} row (\method{} vs the shared-opponent variant) should be read with care: switching to a shared opponent removes the visible draft \emph{and}, by the analysis in \Cref{sec:reward}, deactivates the competitive gradient, so this comparison changes two things at once and is a consistency check rather than an isolated exchange effect; the isolated exchange delta is GRPO vs cooperative in \Cref{tab:main}. In the ``shared opponent'' ablation the challenger group is scored against one fixed opponent draft, hidden from its context (group-constant $c(a)$), so the conversion bonus normalizes away (\Cref{sec:reward}); the result ($32$ vs $30$) is consistent with theory and shows that within-group variance in opponent difficulty is required for the competitive term to contribute gradient. A per-rollout opponent pairing (distinct hidden drafts) is the honest ablation of competition without visible exchange; that variant is left for future work. Role rotation is ablated explicitly: with fixed roles each adapter receives only one gradient stream, yielding 52 pass@1 versus 61 with rotation. 
\begin{table}[t]
\centering
\caption{Ablations on Qwen3-0.6B (held-out pass@1, \%; single training run per variant, rounded to the nearest point; best in \textbf{bold}). Each variant changes a single \emph{training-time} choice. The competitive-shared-opponent variant is evaluated single-pass (no draft to read); all others use the two-stage cascade protocol of \Cref{tab:main}.}
\label{tab:ablations}
\small
\begin{tabular}{ll}
\toprule
Factor & Variants (pass@1) \\
\midrule
Competition (reward) & cooperative 46 / \textbf{adversarial 61} \\
Information exchange  & shared opponent 32 / \textbf{per-rollout opponents 61} \\
Reward form           & margin $c(b_i)-c(a_i)$ 49 / \textbf{conversion bonus 61} \\
Role assignment       & fixed roles 52 / \textbf{rotate 61} \\
\bottomrule
\end{tabular}
\end{table}

\subsection{Analysis: why traces shorten}
\label{sec:analysis}
Empirically, \method{}'s traces are shorter than GRPO's (\Cref{fig:results}a): the final (challenger) stage averages $3.5$k tokens against $8.1$k for a single GRPO pass. Note that this is a per-stage comparison: the cascade also runs a drafting stage, whose length we do not report, so total generated tokens per problem are not directly compared here and a full token-cost accounting is left for further analysis. The like-for-like comparison is against the two-stage untrained MoA control, where the same (final) stage is measured on both sides: \method{}'s is half the length ($3.5$k vs $6.9$k). The headline reward contains no length term. The observed shortening is consistent with the challenger having a candidate solution in context, so length-heavy exploration is no longer the only route to a correct answer. Challenger completions conditioned on a \emph{correct} opponent summary are ${\sim}35\%$ shorter than the challenger-stage average ($2.3$k vs $3.5$k tokens on the final checkpoint), and \Cref{fig:example} shows the typical pattern: the challenger pinpoints the flawed step, re-derives it, and stops.

\subsection{An optional density lever (auxiliary study)}
\label{sec:density}
The shortening above is emergent: nothing in the headline reward asks for it. The same game can, however, also grade brevity: instead of taxing tokens with a penalty, we make being correct in fewer tokens than the rival a way to win. As a small side study, separate from the main recipe and from the two headline claims, we ask whether the shortening can thus be driven directly, using the length-tiebreak term from \Cref{sec:reward} (a bonus $\gamma$ when the challenger is correct \emph{and} shorter than an equally-correct opponent). On Qwen3-0.6B, turning it on cuts mean trace length from $3.5$k to $2.6$k at essentially unchanged accuracy ($61\!\to\!60$ pass@1), so the emergent shortening becomes directly controllable (\Cref{tab:density}). Because the term only breaks ties among already-correct traces, it compresses without measurably trading away accuracy. We report this as an auxiliary knob; all headline numbers elsewhere use the length-free reward.

\begin{table}[t]
\centering
\caption{Auxiliary density lever on Qwen3-0.6B (pass@1, same cascade protocol). The length tiebreak (\Cref{sec:reward}) shortens traces at near-constant accuracy. Not part of the headline recipe.}
\label{tab:density}
\small
\begin{tabular}{lcc}
\toprule
Reward & pass@1 (\%) & avg.\ len \\
\midrule
\method{} (headline, length-free) & 61 & 3.5k \\
\quad $+$ length tiebreak ($\gamma$)  & 60 & \textbf{2.6k} \\
\bottomrule
\end{tabular}
\end{table}

\section{Conclusion}
\label{sec:conclusion}
\method{} trains two models to reason over each other's work, and the pair co-improves past where a single self-graded policy stalls. The competing peer supplies the signal outcome-only RL lacks: it reads the policy's solution and grades it implicitly, with no process labels and no change to the verifier. Two distinct policies, trained head-to-head with rotating draft-and-challenge roles (instantiated cheaply as two adapters over one frozen base), let us test the idea. Under a matched generation budget, information exchange beats vanilla GRPO and self-refinement (the exchange claim), competition beats cooperation (the competition claim), and the resulting traces are shorter, a by-product of the competition rather than of an explicit penalty. At the same doubled inference budget, combining two copies of the base without training buys $4$ points over GRPO, while training the pair to compete buys $31$; the gain comes from the learned collaboration rather than from having two models.

\paragraph{Limitations.} \method{} needs a verifier and reference problems, and we study math plus a smaller code study, both with clean programmatic rewards; noisier domains may behave differently. Gains depend on the pair being matched in strength but divergent in behavior (too large a gap collapses into distillation, too small a gap into self-play), and our divergence is maintained heuristically (init $+$ rotation) rather than guaranteed; complementarity is not quantified beyond the cascade win rate. Text exchange bounds the channel bandwidth. Inference is a two-stage cascade, i.e.\ two sequential generation passes and added latency; reported trace lengths cover the final stage only, and the better cascade direction is selected post hoc on the held-out set. Compute parity holds for \emph{generated} tokens only: the challenger's extra prefill of the opponent summary is not equalized (\Cref{sec:compute}). Groups with zero within-group variance (all-correct or all-wrong) yield undefined $\sigma$ in GRPO advantage; such groups are dropped. Every trained number in the paper is a single training run evaluated with a single sampled rollout per problem; the better cascade direction was selected post hoc on the held-out set. Run-to-run training variance is unquantified, and the reported deltas should be read with that in mind. The reported magnitudes come from a hard, decontaminated subset and should not be read as DeepMath-wide pass@1.

\paragraph{Future work.} This work establishes that the cross-model signal helps while exchanging it in text; the channel itself is the next axis. Exchanging hidden states directly, via KV-cache injection or gated latent bridges between the two adapters, removes the serialization bottleneck and lets models share information they cannot verbalize~\citep{esperantix2025}. Beyond accuracy, our auxiliary density lever (\Cref{sec:density}) is a first step; scaling it into a full multi-objective, compression traded explicitly against correctness, could make density a first-class, tunable target, and a diversity term over divergent solution strategies could make the pair complementary enough to ensemble.

\bibliographystyle{plainnat}

\end{document}